\documentclass[10pt, a4paper]{article}
\usepackage{lrec2022} 
\usepackage{multibib}
\newcites{languageresource}{Language Resources}
\usepackage{graphicx}
\usepackage{tabularx}
\usepackage{amsmath}
\usepackage{soul}
\usepackage{hyperref}
\usepackage[noabbrev,capitalize,nameinlink]{cleveref}
\usepackage{booktabs}
\usepackage{tikz}
\usepackage{tikz-dependency}
\usepackage{xcolor, soul, colortbl}
\usepackage{array, multirow}
\usepackage{adjustbox}
\usepackage{algorithm}
\usepackage{algpseudocode}
\usepackage[shortlabels]{enumitem}
\usepackage{tabularx}
\usepackage{amssymb}
\usepackage{pifont}
\newcommand{\cmark}{\ding{51}}%
\newcommand{\xmark}{\ding{55}}%

\usepackage{titlesec}
\titleformat{\section}{\normalfont\large\bfseries\center}{\thesection.}{1em}{}
\titleformat{\subsection}{\normalfont\SmallTitleFont\bfseries\raggedright}{\thesubsection.}{1em}{}
\titleformat{\subsubsection}{\normalfont\normalsize\bfseries\raggedright}{\thesubsubsection.}{1em}{}
\renewcommand\thesection{\arabic{section}}
\renewcommand\thesubsection{\thesection.\arabic{subsection}}
\renewcommand\thesubsubsection{\thesubsection.\arabic{subsubsection}}

\newcommand{\std}[2]{{#1}{\footnotesize$\pm${#2}}}
\newcommand{\bertb}{BERT\textsubscript{base}}
\newcommand*\circled[1]{\tikz[baseline=(char.base)]{
            \node[shape=circle,draw,inner sep=.6pt] (char) {#1};}}

\usepackage{epstopdf}
\usepackage[utf8]{inputenc}

\usepackage{hyperref}
\usepackage{xstring}

\usepackage{color}

\title{\textsc{Kompetencer}: Fine-grained Skill Classification in Danish Job Postings via Distant Supervision and Transfer Learning}

\name{Mike Zhang\textsuperscript{*}\textsuperscript{$\diamondsuit$} \hspace{1em} Kristian Nørgaard Jensen\textsuperscript{*}\textsuperscript{$\diamondsuit$} \hspace{1em} Barbara Plank\textsuperscript{$\diamondsuit$}\textsuperscript{$\clubsuit$}} 

\address{
\textsuperscript{$\diamondsuit$}Department of Computer Science, IT University of Copenhagen, Denmark \\
\textsuperscript{$\clubsuit$}Center for Information and Language Processing (CIS), LMU Munich, Germany\\
\texttt{\{mikz, krnj\}@itu.dk} \hspace{2em} {\tt bplank@cis.uni-muenchen.de}}

\abstract{
Skill Classification (SC) is the task of classifying job competences from job postings. This work is the first in SC applied to Danish job vacancy data. We release the first Danish job posting dataset: \textsc{Kompetencer} (\emph{en}: competences), annotated for nested spans of competences. To improve upon coarse-grained annotations, we make use of The European Skills, Competences, Qualifications and Occupations (ESCO;~\newcite{le2014esco}) taxonomy API to obtain fine-grained labels via distant supervision. We study two setups: The zero-shot and few-shot classification setting. We fine-tune English-based models and RemBERT~\cite{chung2020rethinking} and compare them to in-language Danish models. Our results show RemBERT significantly outperforms all other models in both the zero-shot and the few-shot setting.
 \\ \newline \Keywords{Skill Classification, Distant Supervision, Transfer Learning, Domain Adaptive Pretraining, Job Postings}}

\begin{document}

\maketitleabstract
\begingroup
\renewcommand\thefootnote{*}
\footnotetext{The authors contributed equally to this work.}
\endgroup

\section{Introduction}

Job Posting data (JPs) is emerging on a variety of platforms in big quantities, and can provide insights on labor market skill set demands and aid job matching~\cite{balog2012expertise}. \textit{Skill Classification} (SC) is to classify competences (i.e., hard and soft skills) necessary for any occupation from unstructured text or JPs. 

Several works focus on Skill Identification~\cite{jia2018representation,sayfullina2018learning,tamburri2020dataops}. This is to classify whether a skill occurs in a sentence or job description. However, continuing the pipeline, there is little work in further categorizing the identified skills by leveraging taxonomies such as ESCO. Another limitation is the scope of language, where all previous work focus on English job postings. This hinders in particular local job seekers from finding an occupation suitable to their specific skills within their community via online job platforms. 

\begin{figure}[t]
    \centering
    \includegraphics[width=.94\linewidth]{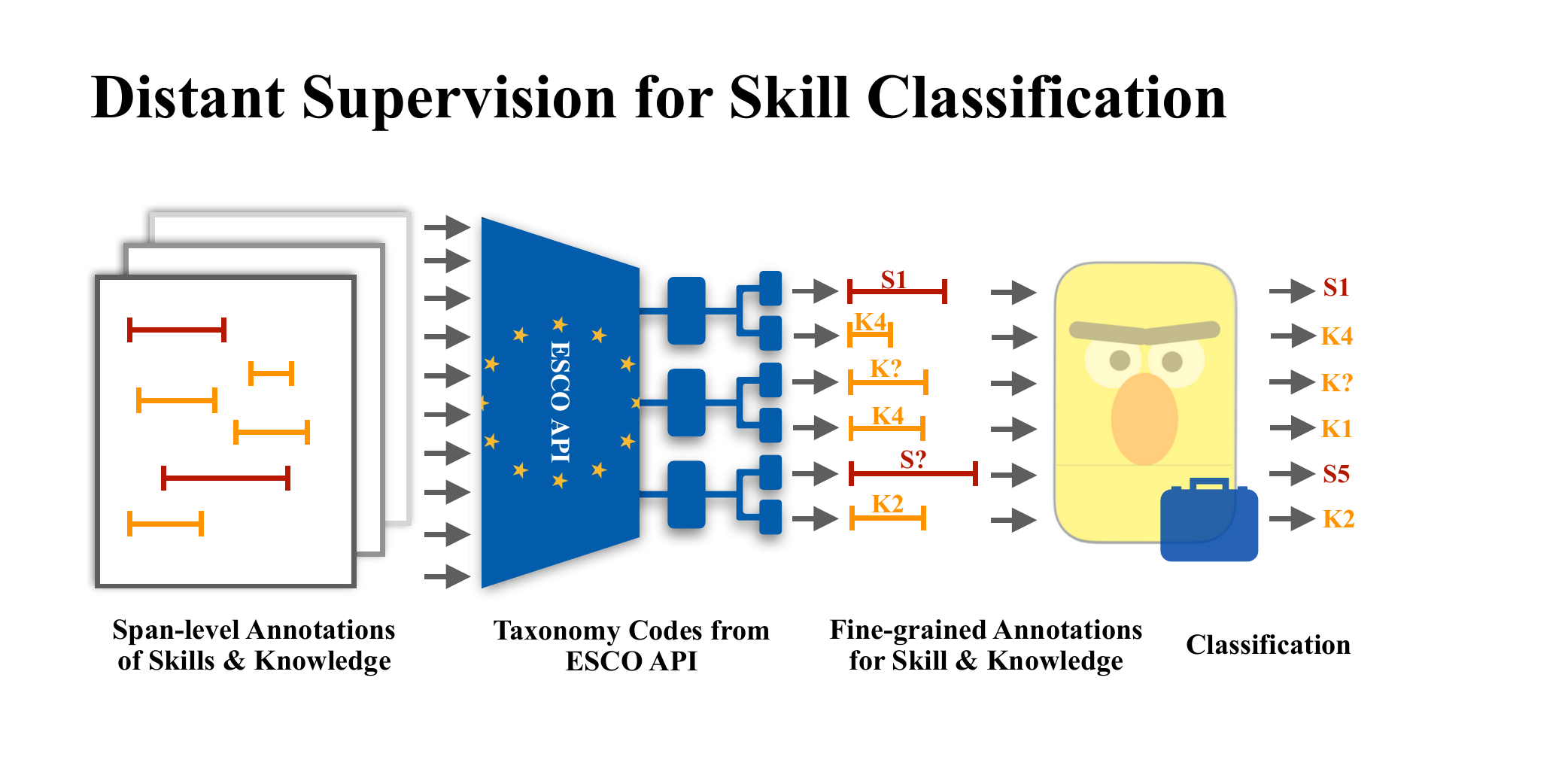}
    \looseness=-1
    \caption{\textbf{Pipeline for Fine-grained Danish Skill Classification.} We propose a distant supervision pipeline, where we have identified spans of skills and knowledge. We query the ESCO API and fine-tune a model on the distantly supervised labels.}
    \label{fig:pipeline}
\end{figure}

In this work, we look into the Danish labor market. We introduce \textsc{Kompetencer}, a novel Danish job posting dataset annotated on the \emph{span-level} for nested \emph{Skill} and \emph{Knowledge} Components (SKCs) in job postings. We do not directly annotate for the fine-grained taxonomy codes from e.g., ESCO, but rather annotate more generic spans of SKCs (\cref{fig:doccano}), and then exploit the ESCO API to bootstrap fine-grained SKCs via distant supervision~\cite{mintz2009distant} and create ``silver'' data for skill classification. Our proposed distant supervision pipeline is denoted in~\cref{fig:pipeline}.

Recently, Natural Language Processing has seen a surge of several transfer learning methods and architecture which help improve state-of-the-art significantly on several tasks~\cite{peters2018deep,howard2018universal,radford2018improving,devlin2019bert}.
In this work, we explore the benefits of zero-shot cross-lingual transfer learning with English \bertb{}~\cite{devlin2019bert} and a \bertb{} that we continuously pretrain~\cite{han-eisenstein-2019-unsupervised,gururangan2020don} on 3.2M English JP sentences and test on Danish and compare it to in-language models: Danish BERT and our domain-adapted Danish BERT model on 24.5M Danish JP sentences. We analyze the zero-shot transfer of English to Danish SC. Last, we experiment with few-shot training: We fine-tune a multilingual model~\cite{chung2020rethinking} on English JPs with a few Danish JPs and show how zero-shot transfer compares to training on a small amount of in-language data.

\paragraph{Contributions}\circled{1} We release \textsc{Kompetencer},\footnote{\url{https://github.com/jjzha/kompetencer}} the first Danish Skill Classification dataset with distantly supervised fine-grained labels using the ESCO taxonomy. \circled{2} We furthermore present experiments and analysis with in-language Danish models vs.\ a zero-shot cross-lingual transfer from English to Danish with domain-adapted BERT models. \circled{3} We target a few-shot learning setting with a multilingual model trained on both English and a few Danish JPs.

\section{\textsc{Kompetencer} Dataset}\label{sec:data}

\begin{table}[t]
    \centering
    \resizebox{.88\linewidth}{!}{
    \begin{tabular}{l|rr}
    \toprule
    $\downarrow$ \textbf{Statistics}, \textbf{Language} $\rightarrow$ & \textsc{\textbf{English (EN)}} & \textsc{\textbf{Danish (DA)}} \\
    \midrule
    \textbf{\# Posts}                   & 391      &  60    \\
    \textbf{\# Sentences}               & 14,538   &  1,479 \\
    \textbf{\# Tokens}                  & 232,220  &  20,369\\
    \textbf{\# Skill Spans}             & 6,576    &  665   \\
    \textbf{\# Knowledge Spans}         & 6,053    &  255   \\
    \midrule
    $\mathbf{\bar{x}}$ \textbf{Skill Span}        & 3.97  & 3.71\\
    $\mathbf{\bar{x}}$ \textbf{Knowledge Span}    & 1.80  & 1.73\\
    $\mathbf{\tilde{x}}$ \textbf{Skill Span}      & 4     & 3   \\
    $\mathbf{\tilde{x}}$ \textbf{Knowledge Span}  & 1     & 1   \\
    \textbf{Skill [90\%]}                         & [1, 9]& [1, 9]    \\
    \textbf{Knowledge [90\%]}                     & [1, 5]& [1, 4]    \\
    \midrule
    \textbf{Silver fine-grained labels}                      & \cmark & \xmark \\
    \textbf{Gold fine-grained labels}                        & \xmark & \cmark \\
    \bottomrule
    \end{tabular}}
    \caption{\textbf{Statistics of Annotated Dataset.} We report the total number of JPs across languages and their respective number of sentences, tokens, and SKCs. Below, we show the mean length of SKCs ($\mathbf{\bar{x}}$), median length of SKCs ($\mathbf{\tilde{x}}$), and the 90th percentile of length [90\%] starting from length 1. We also indicate the type of labels in both sets (silver or gold labels). The EN set is larger than the DA split.}
    \label{tab:num-posts}
\end{table}

\subsection{Skill \& Knowledge Definition}
There is an abundance of competences and there have been large efforts to categorize them. 
The European Skills, Competences, Qualifications and Occupations (ESCO;~\newcite{le2014esco}) taxonomy is the standard terminology linking skills, competences and qualifications to occupations. The ESCO taxonomy mentions three categories of competences: \emph{Knowledge}, \emph{skill}, and \emph{attitudes}. ESCO defines knowledge as follows:
\begin{quote}
    ``Knowledge means the outcome of the assimilation of information through learning. Knowledge is the body of facts, principles, theories and practices that is related to a field of work or study.''~\footnote{\scriptsize{\href{https://ec.europa.eu/esco/portal/escopedia/Knowledge}{\url{ec.europa.eu/esco/portal/escopedia/Knowledge}}}}
\end{quote}
For example, a person can acquire the Python programming language through learning. This is denoted as a \emph{knowledge} component and can be considered generally a \emph{hard skill}. However, one also needs to be able to apply the knowledge component to a certain task. This is known as a \emph{skill} component. ESCO formulates it as:
\begin{quote}
    ``Skill means the ability to apply knowledge and use know-how to complete tasks and solve problems.''~\footnote{\scriptsize{\href{https://ec.europa.eu/esco/portal/escopedia/Skill}{\url{ec.europa.eu/esco/portal/escopedia/Skill}}}}
\end{quote}
 
\noindent
In ESCO, the \emph{soft skills} are referred to as \emph{attitudes}. ESCO considers attitudes as skill components:

\begin{quote}
    ``The ability to use knowledge, skills and personal, social and/or methodological abilities, in work or study situations and professional and personal development.''~\footnote{\scriptsize{\href{http://data.europa.eu/esco/skill/A}{\url{data.europa.eu/esco/skill/A}}}}
\end{quote}

\noindent
To sum up, hard skills are usually referred to as \emph{knowledge} components, and applying these hard skills to something is considered a \emph{skill}. Then, soft skills are referred to as \emph{attitudes}, these are part of skill components. There has been no work, to the best of our knowledge, in annotating skill and knowledge components in JPs.

\begin{figure}[t]
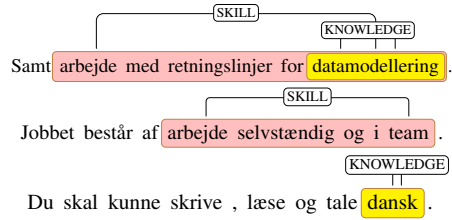

\centering
\resizebox{.8\linewidth}{!}{
\begin{dependency}[edge slant=0pt, edge vertical padding=3pt, edge style={-}]
    \begin{deptext}
    Samt \& arbejde \&  med \&  retningslinjer \&  for \&  datamodellering \& . \\
    \end{deptext}
    \wordgroup[group style={fill=pink, draw=brown, inner sep=.3ex}]{1}{2}{6}{sk}
    \wordgroup[group style={fill=yellow, draw=brown, inner sep=.1ex}]{1}{6}{6}{kn}
    \depedge[edge height=4ex]{2}{6}{SKILL}
    \depedge[edge height=2ex, edge horizontal padding=-18pt, edge end x offset=10pt]{6}{6}{KNOWLEDGE}
\end{dependency}}

\resizebox{.77\linewidth}{!}{
\begin{dependency}[edge slant=0pt, edge vertical padding=2pt, edge style={-}]
    \begin{deptext}
    Jobbet \& består \& af \& arbejde \& selvstændig \& og \& i \& team \&.\\
    \end{deptext}
    \wordgroup[group style={fill=pink, draw=brown, inner sep=.1ex}]{1}{4}{8}{sk}
    \depedge[edge height=2ex]{4}{8}{SKILL}
\end{dependency}}

\resizebox{.75\linewidth}{!}{
\begin{dependency}[edge slant=0pt, edge vertical padding=2pt, edge style={-}]
    \begin{deptext}
    Du \& skal \& kunne \& skrive \& , \& læse \& og \& tale \& dansk \& .\\
    \end{deptext}
    \wordgroup[group style={fill=yellow, draw=brown, inner sep=.1ex}]{1}{9}{9}{kn}
    \depedge[edge height=2ex]{9}{9}{KNOWLEDGE}
\end{dependency}}

    \caption{\textbf{Examples of Skills and Knowledge Components.} Annotated samples of passages in varying Danish job postings. SKCs can be nested as shown in the first example.}
    \label{fig:doccano}
\end{figure}

\subsection{Dataset Statistics}
Both the English and Danish data comes from a large job platform with various types of JPs.\footnote{We release the annotated spans in~\url{https://github.com/jjzha/kompetencer/tree/master/data}} The English JPs are from~\newcite{zhang2022skillspan}.  In~\cref{tab:num-posts}, we show the statistics of both the annotated English and Danish data split. We note that the number of English JPs is larger than the Danish split. For Danish, there are fewer knowledge spans proportional to English. Apart from this, both the English and Danish JPs follow a similar trend in terms of statistics. The mean length of skills and knowledge ($\mathbf{\bar{x}}$) is slightly shorter for Danish, 3.97 vs. 3.71 and 1.80 vs. 1.73 respectively. The median length of skills ($\mathbf{\tilde{x}}$) is one token shorter for Danish. However, we note again that the length of skills can vary substantially, ranging from 1--9 for both languages. Then, for knowledge components this ranges from 1--5 and 1--4 for English and Danish respectively. The similarity in statistics shows the consistency of annotations, which we elaborate on in the next section.

\cref{fig:doccano} shows some examples of the annotated SKCs. ``Samt arbejde med retningslinjer for datamodellering'' (\emph{en}: ``As well as working with guidelines for data modeling''), shows a nesting example: ``datamodellering'' shows a knowledge component (i.e., something that one can learn), and the skill is to apply it. ``Jobbet består af arbejde selvstændig og i team'' (\emph{en}: The job consists of working independently and in a team) indicates an \emph{attitude} as ``working independently or in a team'' is a social ability. We furthermore consider languages a knowledge component, as one can acquire the language through schooling. Overall, the classification of the spans could be a short sentence (i.e., $\leq$9 tokens) or single token classification.

\begin{figure*}[ht]
    \centering
    \includegraphics[width=.81\linewidth]{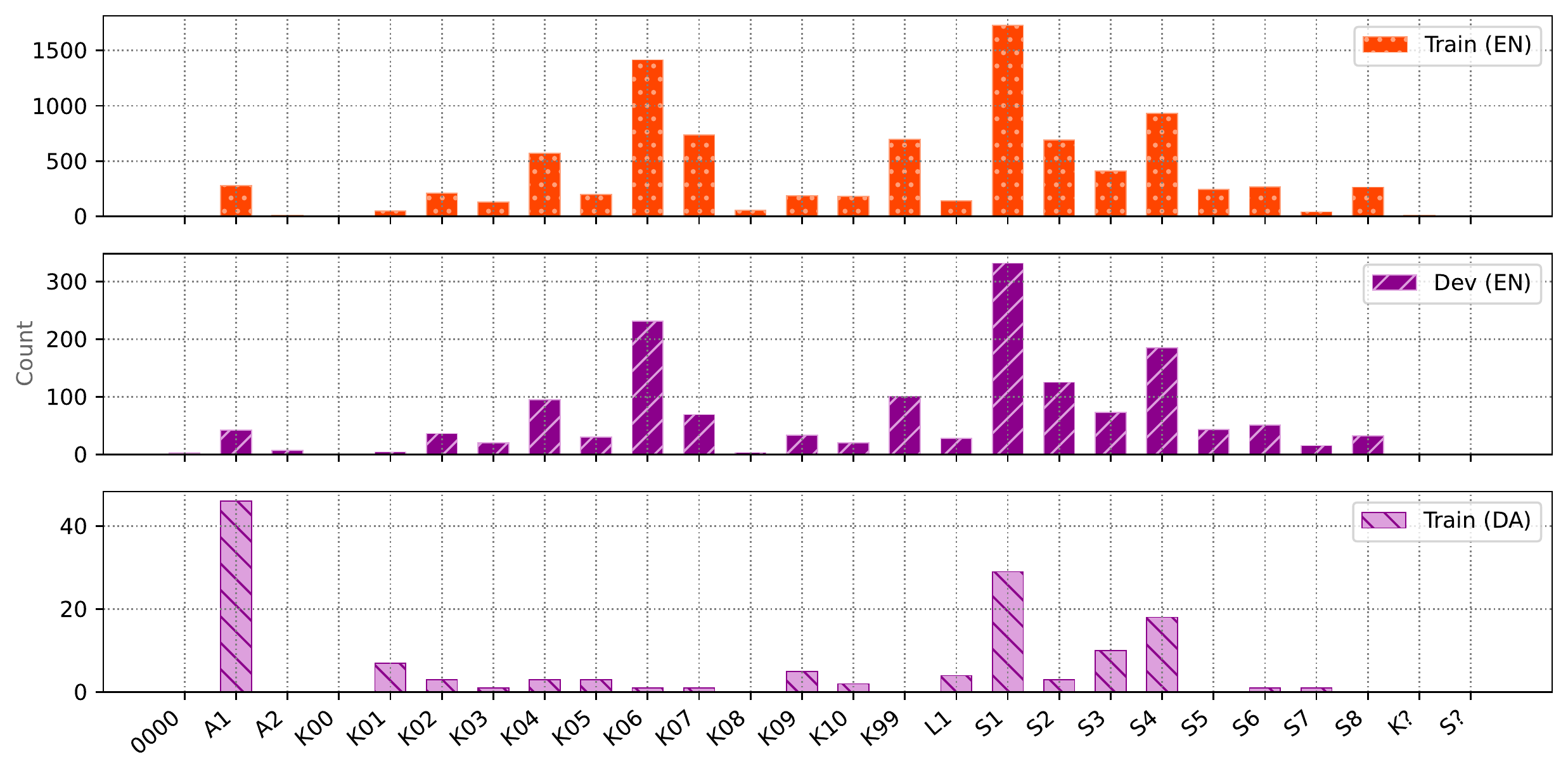}
    \caption{\textbf{Label Distribution of Distantly Supervised Labels.} In the top and middle barplot we show the fine-grained label distribution of the English training and development split respectively. The splits follow a similar distribution. For the Danish training split on the bottom, there is a large increase of \texttt{A1} labels, which indicate more \emph{attitude}-like skills. All splits have a larger fraction of the label \texttt{S1}, which encapsulates communicative skills. Explanations of labels are given in~\cref{tab:lab} (\cref{labelmeaning}, Appendix).}
    \label{fig:distribution}
\end{figure*}
\subsection{Annotations}

\paragraph{Skill Identification Annotations} We annotate with the annotation guidelines denoted in~\newcite{zhang2022skillspan} used on the English data split to identify the SKCs in a JP. There are around 57.5K tokens (approximately 4.6K sentences, in 101 job posts) that was used to calculated agreement on. 
The annotations were compared using Cohen's $\kappa$~\cite{fleiss1973equivalence} between pairs of annotators, and Fleiss' $\kappa$~\cite{fleiss1971measuring}, which generalizes Cohen's $\kappa$ to more than two concurrent annotations. We consider two levels of $\kappa$ calculations: \textbf{\textsc{Token}} is calculated on the token level, comparing the agreement of annotators on each token (including non-entities) in the annotated dataset.~\textbf{\textsc{Span}} refers to the agreement between annotators on the exact span match over the surface string, regardless of the type of named entity, i.e., we only check the position of tag without regarding the type of the named entity. The observed agreement scores over the three annotators is between 0.70--0.75 Fleiss' $\kappa$ for both levels of calculation which is considered a \emph{substantial agreement}~\cite{landis1977measurement}. Then, for the Danish data split, we use the same guidelines as for English. Here, we consider one annotator that annotates for the SKCs.

\begin{algorithm}[ht]
\caption{Getting the best match for a skill in the ESCO API using Levenshtein distance}\label{alg:cap}
\begin{algorithmic}
\Procedure{FetchSkill}{$Skill,Type$}\Comment{Find Skill in the ESCO API}
\State $X \gets \text{Top-100} \ \text{query results from ESCO}$
\State $X \gets \{\texttt{typeof(}x\texttt{)} = Type : x \in X\}$
\State $d \gets \infty$
\State $r \gets \texttt{None}$
\For{$x \in X$}
\State $D \gets \texttt{levenshtein(x, Skill)}$
\If{$D = 0$}
    \State \textbf{return} $x$ \Comment{Perfect match}
\ElsIf{$D < d$}
    \State $r \gets x$
    \State $d \gets D$
\EndIf
\EndFor
\State \textbf{return} $r$ \Comment{Best match based on Levenshtein distance}
\EndProcedure
\end{algorithmic}
\end{algorithm}

\begin{figure*}[ht]
    \centering
    \includegraphics[width=\linewidth]{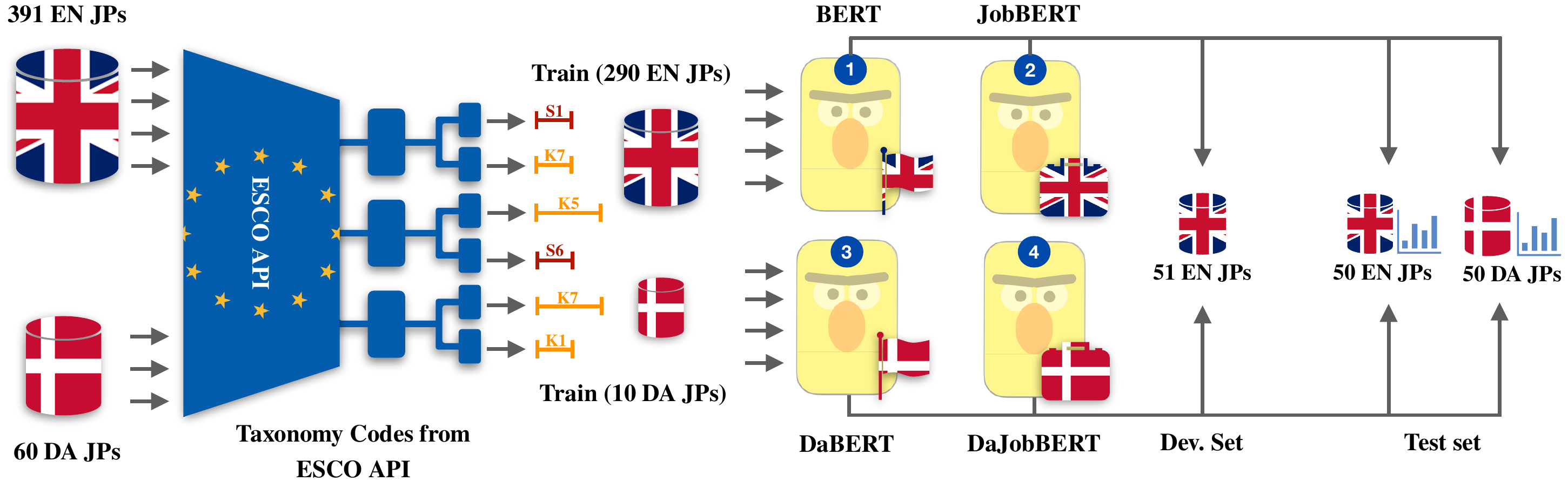}
    \caption{\textbf{Experimental Pipeline for Fine-grained Danish Skill Classification.} Read from left to right, we start with each respective dataset for English (EN) and Danish (DA). We obtain the labels from the ESCO API and train for each language split two models: For EN, these are (1) \bertb{} and (2) JobBERT. For DA, these are (3) DaBERT and (4) DaJobBERT. The Danish data is split into 10/50 train/test and the English data in to 290/51/50 train/dev/test JPs. The Danish models are fine-tuned on the Danish train set, and use \emph{no} in-language development set (i.e., English dev.). In the end all models are applied to the Danish and English test set separately.}
    \label{fig:experiment-pipeline}
\end{figure*}

\paragraph{Fine-grained Annotations}\label{sec:fine-anno} Currently, our proposed dataset consists of identified SKCs. To obtain fine-grained labels of each span, we explore distant supervision using the ESCO API, where the setup is broadly depicted in~\cref{fig:pipeline}. The annotated spans are queried to the API, then via~\cref{alg:cap}, we determine whether the obtained SKC is ``relevant'' or not via Levenshtein distance matching~\cite{levenshtein1966binary}. In addition, we determine the quality of the distant supervised labels by human evaluation. We manually check each of the annotated spans to its obtained label from the ESCO API. After checking a subset 2,622 English labels --- without correcting --- and its distantly supervised labels, we obtain 41.3\% accuracy on the correctness of the distantly supervised labels. We note that across all 9,473 labels in the original English training and development data (details of train/dev/test splits in~\cref{sec:expsetup}), a total of 7.4\% is unidentified by the ESCO database, and is thus labeled by \texttt{K99} from ESCO in the resulting train and development data here. For the Danish data, we obtain 70.4\% accuracy on the training set and 20.2\% is missing, albeit the Danish training set only contains 138 SKCs. For the Danish test set, we correct the distantly supervised labels to create a gold test set. Here, 14.1\% was initially correct and 23.5\% missed a label. In~\cref{fig:distribution}, we show the distantly supervised fine-grained label distribution of the English training and development set split, and the Danish training split. The following labels: \texttt{0000}, \texttt{K?}, and \texttt{S?} are artifacts of querying the ESCO API (i.e., unidentified skills). We did not employ any post-processing and left them as is. We presumed they would not influence the model significantly as their numbers are low.



\section{Methodology}\label{sec:method}
In the current setting, we have annotated spans of SKCs. We extract the spans from the JPs and query the ESCO API to obtain silver labels. We formulate this task as a text classification problem. We consider a set of JPs $\mathcal{D}$, where $d \in \mathcal{D}$ is a set of extracted spans (\emph{and not full sentences}) with the $i^\text{th}$ span $\mathcal{X}^i_{d} = \{x_1, x_2, ..., x_T\}$ and a target class $c \in \mathcal{C}$, where $\mathcal{C} = \{\text{\texttt{S*}}, \text{\texttt{K*}}\}$. The labels \texttt{S*} and \texttt{K*} depend on the distantly supervised ESCO taxonomy code (e.g., S4: Management Skills,\footnote{\url{http://data.europa.eu/esco/skill/S4}} K2: Arts and Humanities\footnote{\url{http://data.europa.eu/esco/isced-f/02}}). The goal of this task is then to use $\mathcal{D}$ to train an algorithm $h: \mathcal{X} \mapsto \mathcal{C}$ to accurately predict skill tags by assigning an output label $c$ for input $\mathcal{X}^i_{d}$.

\subsection{Encoders}
As baseline for Danish SC, we consider a Danish BERT (\textbf{DaBERT}) encoder.\footnote{\url{https://huggingface.co/Maltehb/danish-bert-botxo}} Following~\newcite{gururangan2020don}, we continuously pretrain DaBERT on 24.5M Danish JP sentences for \emph{one} epoch, we name this \textbf{DaJobBERT}.\footnote{\url{https://huggingface.co/jjzha/dajobbert-base-cased}}
To test zero-shot performance from English to Danish for SC, we use \textbf{\bertb{}}~\cite{devlin2019bert} and a domain-adapted \bertb{} model on 3.2M JP sentences, namely \textbf{JobBERT}~\cite{zhang2022skillspan}. We assume that domain-adapted models like JobBERT and DaJobBERT would improve SC as the ``domain'' is the same.

\paragraph{Multilingual Encoder}
We explore whether using a multilingual encoder would benefit the classification of skills for Danish in a low-resource setting. For the experiments we use \textbf{RemBERT}~\cite{chung2020rethinking}, it has recently shown to outperform mBERT~\cite{devlin2019bert} on several tasks. All models are using a final Softmax layer for the classification of spans.

\subsection{Experimental Setup}\label{sec:expsetup}
Our detailed experimental setup is shown in~\cref{fig:experiment-pipeline}. We start with 391 English and 60 Danish job postings (\cref{tab:num-posts}) annotated with spans of SKCs. The spans are then queried to the ESCO API (\cref{fig:pipeline}). We split the English data into 290 train (9,472 SKCs), 51 dev (1,577 SKCs), and 50 JPs for test (1,578 SKCs), and for the Danish data we split this into 10 JPs (138 SKCs) for training and 50 JPs for test (782 SKCs). For the label distribution we refer back to~\cref{fig:distribution} (excl. test).

\begin{figure*}[ht]
    \centering
    \includegraphics[width=.85\linewidth]{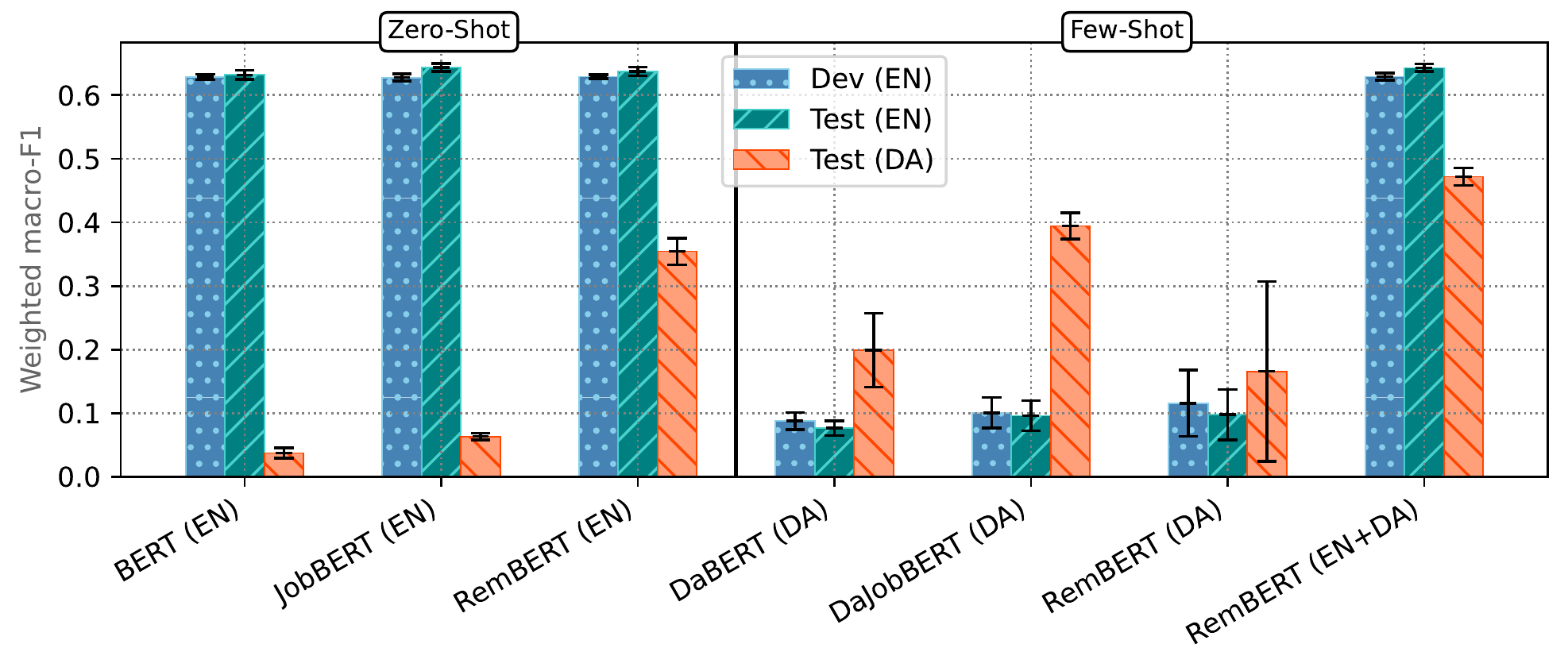}
    \looseness=-1
    \caption{\textbf{Performance of Models on English and Danish.} We test seven setups on several splits of data: English development (\textsc{\textbf{Dev (EN)}}), English test set (\textsc{\textbf{Test (EN)}}), and Danish test set (\textsc{\textbf{Test (DA)}}). Reported is the weighted macro-F1. The whiskers indicate each respective standard deviation of runs on five random seeds. Left side of the black vertical line indicates a full zero-shot setting on \textsc{Test (DA)}, on the right shows the few-shot setting on the same test set. With respect to the models, language abbreviation in brackets (e.g., \textbf{BERT (EN)}) indicates what it has been fine-tuned on. Exact numbers including significance testing are noted in~\cref{tab:exactresults} (\cref{app:results}, Appendix).}
    \label{fig:results}
\end{figure*}

We fine-tune \bertb{} and JobBERT on the spans in 290 English JPs. Next, we fine-tune DaBERT and DaJobBERT on the 10 Danish JPs. For RemBERT, we fine-tune in three ways: Only on English, only on Danish, and on both English and Danish together.
For all setups, we choose the model with the best score on the English dev.\ set. As pointed out by~\newcite{artetxe2020call}: Pure unsupervised cross-lingual transfer should not use any cross-lingual signal by definition. As our attention is on Danish, we do not use any Danish labeled training data \emph{nor} dev.\ data in the zero-shot setting. All models in the end will be tested on the held-out 50 English and Danish JPs separately.\footnote{The English test set contains silver labels (distantly supervised), while the Danish test set is human corrected (gold).} In summary, we have three setups: (1) Fine-tuned on English JPs only (BERT, JobBERT, RemBERT), (2) fine-tuned on Danish JPs only (DaBERT, DaJobBERT, RemBERT), and (3) fine-tuned on both English and Danish JPs (RemBERT). We consider (1) a zero-shot setting, while (2) and (3) do have access to some Danish training data, hence this is a few-shot setting. Throughout the experiments, we use the \textsc{MaChAmp} (v0.3) toolkit~\cite{van-der-goot-etal-2021-massive} for classification. All reported results are the average over five runs with different random seeds on weighted macro-F1.

\section{Analysis of Results}

We show the experimental results in~\cref{fig:results}. Plotted is the weighted macro-F1 of all three setups with seven models and their corresponding standard deviation on the English development set, English test set, and the Danish test set. All models left of the black vertical line are the zero-shot setup, applied to Danish. On the right, these models are in the few-shot setting, this is due to the model having access to some target language training data (DA). 

\paragraph{Performance Zero-shot Setting}
For the models trained on English only (BERT, JobBERT, and RemBERT (EN)) when applied to the English development set, all three models perform similarly. They achieve around 0.63--0.64 weighted macro-F1 with little standard deviation: \bertb{} \std{0.628}{0.004}, JobBERT \std{0.628}{0.006}, and RemBERT (EN) \std{0.629}{0.003} weighted macro-F1. Similarly for the English test set: \bertb{} \std{0.632}{0.007}, JobBERT \std{0.644}{0.006}, and RemBERT (EN) \std{0.637}{0.007} weighted macro-F1, where JobBERT significantly outperforms all other models (details in \cref{app:results}).

It is a tacit that the English-based models perform better than the baseline (DaBERT) on English, both dev.\ and test. 
Conversely, the English-based models perform poorly on the Danish test set: \bertb{} \std{0.038}{0.008} and JobBERT \std{0.063}{0.005} weighted macro-F1. However, given a multilingual encoder (RemBERT) only trained on English, gives a significant gain in zero-shot performance (\std{0.354}{0.021}) with little standard deviation and significantly outperforms the other zero-shot setting models including the target-language baseline (DaBERT). We strongly suspect this is due to Danish being included in the pretraining data of RemBERT.

\paragraph{Performance Few-shot Setting} Apart from RemBERT (EN+DA) having access to English data, all other models fine-tuned on Danish perform poorly on English dev.\ and test. The performance of RemBERT (DA) is slightly better than the best performing Danish-only model DaJobBERT (\std{0.098}{0.040} vs.\std{0.096}{0.024} weighted macro-F1 on English test), where our intuition again goes to the pretraining data.

For DA test, DaBERT is a strong baseline, achieving \std{0.199}{0.058} weighted macro-F1 with little Danish training data. RemBERT (DA) did not result in significant gains having pretrained on multiple languages and another intuition could be that this is a result of negative transfer~\cite{rosenstein2005transfer}. Then, DaJobBERT performs better than DaBERT on the Danish test set: \std{0.395}{0.021} weighted macro-F1. Note that we conducted domain adaptive pretraining from the DaBERT checkpoint on 24.5M Danish JP sentences for one epoch with the Masked Language Modeling objective. This shows that in-language \emph{and} in-domain pretraining is beneficial for this specific task of SC.

\paragraph{Combining Training Data} Last, giving RemBERT all training data (English and Danish) results in substantial improvement over all other models in the zero-shot and few-shot setting alike: \std{0.472}{0.014}, which significantly outperforms all other models on Danish test. Henceforth, it is helpful to have a bit of target-language training data for higher resulting performance.

\paragraph{Is Domain Adaptive Pretraining Worth It?} In light of the results, domain adaptive pretraining shows its benefit for both English and Danish fine-tuning. Specifically for Danish, from the baseline (DaBERT), the improvement is close to 0.2 weighted macro-F1 with DaJobBERT. The domain adaptive pretraining took $\sim$35 hours, using 4 GPUs, to pass once over the unlabeled data (24.5M Danish JP sentences). The largest gain is obtained with combining both English and Danish training data: The improvement is around 0.27 weighted macro-F1. However, the 391 EN and 60 DA JPs took around two months of non-stop annotating. In short, there is a trade-off between continuous pretraining on unlabeled text and annotating: (1) Domain adaptive pretraining gives short-term gains with little costs, but there needs to be enough unlabeled data in the right domain. (2) Annotating extra data results in larger gains long-term, but there is more costs involved.

\begin{figure}[t]
    \centering
        \includegraphics[width=\linewidth]{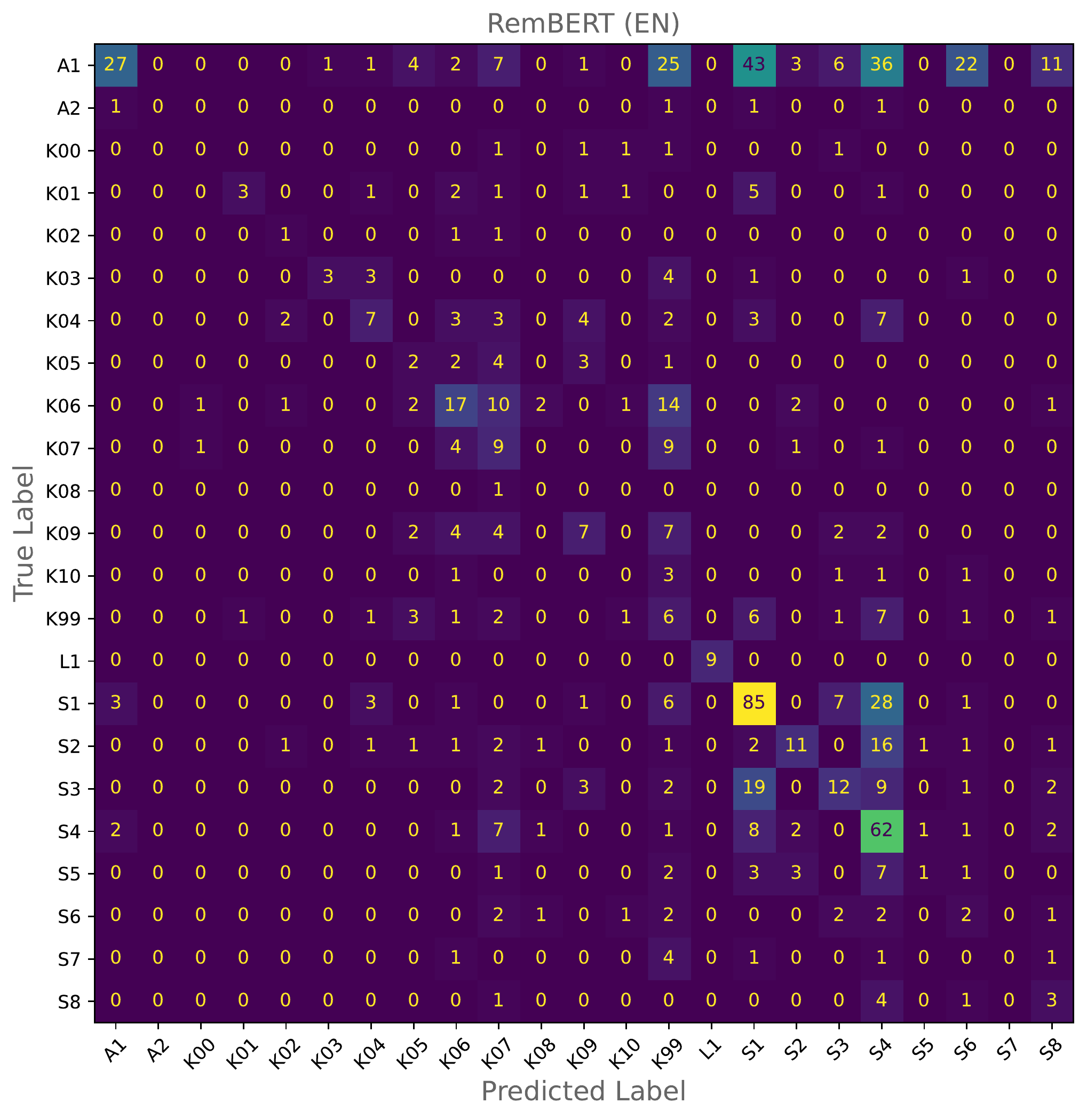}
    \caption{\textbf{Confusion Matrix of RemBERT (EN).} We show the confusion matrix of the zero-shot setting with RemBERT (EN). On the diagonal are the correctly predicted labels. Most of the ``confusion'' is with respect to the labels that encompass the larger fraction of the test set: \texttt{A1}: Attitudes and \texttt{S1}: Communication, collaboration and creativity.}
    \label{fig:cm-en}
\end{figure}

\paragraph{Analysis of Predictions} In~\cref{fig:cm-en}, we show the confusion matrix of the best performing zero-shot model on the test set of the best run and investigate what the model does not predict correctly. In the matrix, the model mostly confuses the label \texttt{A1}, which relates to \emph{attitudes} and gets predicted as \texttt{S1}: Communication, collaboration and creativity. There could be some overlap between these labels as for example the skill ``effektiv'' (\emph{en}: efficient/effective). This is officially labeled as an attitude by ESCO, but a grey area is that ``effective'' could relate to ``creativity''. 

There is also a small cluster of confusion from \texttt{S1-4}. These are rather distinct classes of skills. For example, \texttt{S4} means management skills. A specific example is ``fagligt velfunderet'' (\emph{en}: professionally sound), this could be an attitude. This is all hard to determine since there is no context around the skill. Overall, there is some confusion between the skills when taken out of context. We leave the exploration of fine-grained skill classification \emph{with context} for future work.

\paragraph{Qualitative Analysis Distant Supervision}\label{sec:quality}
We analyze the label selection method and the missing labels in the English dataset as mentioned in \cref{sec:fine-anno}. 
We find that the missing labels in the English data is predominantly coming from technical skills. We found that the missing spans are mostly knowledge components in the form of technologies used today by developers, such as ReactJS, Django, AWS etc.\ This lack of coverage could either be due to specificity or the ever-growing list of technologies. In ESCO, there are several technologies that are listed (e.g., NoSQL, Drupal, WordPress to name a few), but there are also a lot missing (e.g., TensorFlow, Data Science, etc.).

\section{Related Work}

Many works focus on the identification of skills in job descriptions, i.e., whether a sentence contains a skill or not~\cite{sayfullina2018learning,tamburri2020dataops} or what the necessary skills are inferred from an entire job posting~\cite{bhola-etal-2020-retrieving}. We instead identified the SKCs manually in the job descriptions on the sentence-level, as this gives us the highest quality of identified SKCs. Furthermore, there are several works in fine-grained SC (i.e., categorize the skills), but mostly focus on English job descriptions. A straightforward approach is to do exact matching with a predefined list of skills~\cite{malherbe2016bridge,papoutsoglou2017mining,sibarani2017ontology} or do a frequency analysis of skills, cluster them by hand and attach a more general category to them e.g.,~\newcite{gardiner2018skill}. 

Some works have used the ESCO taxonomy directly~\cite{boselli2018classifying,giabelli2020graphlmi}. For example,~\newcite{boselli2018classifying} classified both titles and description for its most suitable ISCO~\cite{elias1997occupational} code (what ESCO is partially based on). However, they only gave one label to each data point (i.e., full job posting), which is unrealistic as most occupations require multiple competences.

Overall, to the best of our knowledge, there seems to be little to no work in directly classifying the identified SKC to a specific ESCO label. In addition, this work is the first of its kind doing this for Danish JPs.

\section{Conclusion}
We present a novel skill classification dataset for competences in Danish: \textsc{Kompetencer}.\footnote{We release the Danish anonymized raw data and annotations of the parts with permissible licenses from a govermental agency which is our collaborator. Links to our English data can be found at \url{https://github.com/kris927b/SkillSpan}. For anonymization, we perform it via manual annotation of job-related sensitive and personal data regarding \texttt{Organization}, \texttt{Location}, \texttt{Contact}, and \texttt{Name} following the work by~\newcite{jensen-etal-2021-de}.} In addition, we transform the coarse-grained human annotated spans to more fine-grained labels via distant supervision with the ESCO API. Our human evaluation shows that the distantly supervised labels give a signal of correctly annotated spans, where we achieve 41.3\% accuracy on a large English label subset, and 70.4\% accuracy on the Danish dev set, and 14.1\% accuracy on the Danish test set. We manually correct the Danish test set with the correct labels from ESCO to create a gold annotated set and keep the English labels as is, and thus silver labels. 

Furthermore, domain adaptive pretraining helps to improve performance on the task specifically for English. The best performance is achieved with RemBERT on both the zero-shot setting (\std{0.354}{0.021} weighted macro-F1) and few-shot setting (\std{0.472}{0.014} weighted macro-F1), where they significantly outperform the other models. The strong performance is likely due to the pretraining data that contains both Danish and English.

Last, since the annotations are on the token-level, this work can be extended to, for example, sequence labeling. We hope this dataset initiates further research in the area of skill classification.







\section{Acknowledgements}
We thank the NLPnorth group for feedback on an earlier version of this paper---in particular Elisa Bassignana and Max M\"uller-Eberstein for insightful discussions. We would also like to thank the anonymous reviewers for their comments to improve this paper. Last, we also thank NVIDIA and the ITU High-performance Computing cluster for computing resources. This research is supported by the Independent Research Fund Denmark (DFF) grant 9131-00019B.









\section{Bibliographical References}\label{reference}

\bibliographystyle{lrec2022-bib}
\bibliography{lrec2022-example}

\section*{Appendix}
\section{Data Statement \textsc{Kompetencer}}\label{app:datastatement}
Following~\newcite{bender-friedman-2018-data}, the following outlines the data statement for \textsc{Kompetencer}:
\begin{enumerate}[A.]
    \itemsep0em
    \item \textsc{Curation Rationale}: Collection of job postings in the English and Danish language for skill classification, to study the impact of skill changes from job postings.
    \item \textsc{Language Variety}: The non-canonical data was collected from the StackOverflow job posting platform, an in-house job posting collection from our national labor agency collaboration partner (\emph{which will be elaborated upon acceptance}), and web extracted job postings from a large job posting platform. US (en-US), British (en-GB) English, and Danish (da-DK) are involved.
    \item \textsc{Speaker Demographic}: Gender, age, race-ethnicity, socioeconomic status are unknown.
    \item \textsc{Annotator Demographic}: Three hired project participants (age range: 25--30), gender: one female and two males, white European and Asian (non-Hispanic). Native language: Danish, Dutch. Socioeconomic status: higher-education students. Female annotator is a professional annotator with a background in Linguistics and the two males with a background in Computer Science.
    \item \textsc{Speech Situation}: Standard American, British English or Danish is used in job postings. Time frame of the data is between 2012--2021.
    \item \textsc{Text Characteristics}: Sentences are from job postings posted on official job vacancy platforms.
    \item \textsc{Recording Quality}: N/A.
    \item \textsc{Other}: N/A.
    \item \textsc{Provenance Appendix}: The Danish job posting data is from our collaborators: The Danish Agency for Labour Market and Recruitment (STAR).
\end{enumerate}

\begin{table}[t]
    \centering
    \resizebox{\linewidth}{!}{
    \begin{tabular}{l|r|r}
    \toprule
    \textsc{\textbf{Parameter}} & \textsc{\textbf{Value}} & \textsc{\textbf{Range}} \\
    \midrule
    Optimizer                           & AdamW                & \\
    $\beta_\text{1}$, $\beta_\text{2}$  & 0.9, 0.99            & \\
    Dropout                             & 0.2                  & 0.1, 0.2, 0.3\\
    Epochs                              & 20                   & \\
    Batch Size                          & 32                   & \\
    Learning Rate (LR)                  & 1e-4                 & 1e-3, 1e-4, 1e-5\\
    LR scheduler                        & Slanted triangular   & \\
    Weight decay                        & 0.01                 & \\
    Decay factor                        & 0.38                 & 0.35, 0.38, 0.5\\
    Cut fraction                        & 0.2                  & 0.1, 0.2, 0.3\\
    \bottomrule
    \end{tabular}}
    \caption{\textbf{Hyperparameters of \textsc{MaChAmp}.}}
    \label{tab:hyperparameters}
\end{table}

\section{Reproducibility}\label{app:hyper}

We use the default hyperparameters in \textsc{MaChAmp}~\cite{van-der-goot-etal-2021-massive} as shown in~\cref{tab:hyperparameters}. For more details we refer to their paper. For the five random seeds we use 3477689, 4213916, 6828303, 8749520, and 9364029. All experiments with \textsc{MaChAmp} were ran on an NVIDIA\textsuperscript{\textregistered} NVIDIA A100-SXM4 40GB GPU
and an AMD\textsuperscript{\textregistered} EPYC 7662 64-Core Processor.

\section{Label Meaning}\label{labelmeaning}
\begin{table*}[t]
    \small
    \centering
    \resizebox{.9\linewidth}{!}{
    \begin{tabularx}{\textwidth}{l|X|X}
    \toprule
    \textsc{\textbf{Label}} & \textsc{\textbf{Subject}} & \textsc{\textbf{Definition}} \\
    \midrule
    0000    & \texttt{ARTIFACT} & \texttt{ARTIFACT}  \\\hline
    A1      & Attitudes & Individual work styles that can affect how well someone performs a job.  \\\hline
    A2      & Values & Principles or standards of behavior, revealing one's judgment of what is important in life.   \\\hline
    K00     & Generic programmes and qualifications & Generic programmes and qualifications are those providing fundamental and personal skills education which cover a broad range of subjects and do not emphasise or specialise in a particular broad or narrow field. \\\hline
    K01     &  Education & \texttt{NO-DEFINITION} \\\hline
    K02     &  Arts and humanities & \texttt{NO-DEFINITION} \\\hline
    K03     &  Social sciences, journalism and information & \texttt{NO-DEFINITION}\\\hline
    K04     &  Business, administration and law & \texttt{NO-DEFINITION} \\\hline
    K05     &  Natural sciences, mathematics and statistics & \texttt{NO-DEFINITION}\\\hline
    K06     &  Information and communication technologies (icts) & \texttt{NO-DEFINITION}\\\hline
    K07     &  Engineering, manufacturing and construction not elsewhere classified & \texttt{NO-DEFINITION}\\\hline
    K08     & Agriculture, forestry, fisheries and veterinary & \texttt{NO-DEFINITION}\\\hline
    K09     & Health and welfare & \texttt{NO-DEFINITION}\\\hline
    K10     & Services & \texttt{NO-DEFINITION}\\\hline
    K99     & Field unknown & \texttt{NO-DEFINITION}\\\hline
    L1      & Languages & Ability to communicate through reading, writing, speaking and listening in the mother tongue and/or in a foreign language. \\\hline
    S1      & Communication, collaboration and creativity &  Communicating, collaborating, liaising, and negotiating with other people, developing solutions to problems, creating plans or specifications for the design of objects and systems, composing text or music, performing to entertain an audience, and imparting knowledge to others.\\\hline
    S2      & Information skills & Collecting, storing, monitoring, and using information; Conducting studies, investigations and tests; maintaining records; managing, evaluating, processing, analysing and monitoring information and projecting outcomes.\\\hline
    S3      & Assisting and caring & Providing assistance, nurturing, care, service and support to people, and ensuring compliance to rules, standards, guidelines or laws.\\\hline
    S4      & Management skills & Managing people, activities, resources, and organisation; developing objectives and strategies, organising work activities, allocating and controlling resources and leading, motivating, recruiting and supervising people and teams.\\\hline
    S5      & Working with computers & Using computers and other digital tools to develop, install and maintain ICT software and infrastructure and to browse, search, filter, organise, store, retrieve, and analyse data, to collaborate and communicate with others, to create and edit new content.\\\hline
    S6      & Handling and moving & Sorting, arranging, moving, transforming, fabricating and cleaning goods and materials by hand or using handheld tools and equipment. Tending plants, crops and animals.\\\hline
    S7      & Constructing & Building, repairing, installing and finishing interior and exterior structures.\\\hline
    S8      & Working with machinery and specialised equipment & Controlling, operating and monitoring vehicles, stationary and mobile machinery and precision instrumentation and equipment. \\\hline
    K?      & \texttt{ARTIFACT} & \texttt{ARTIFACT}  \\\hline
    S?      & \texttt{ARTIFACT} & \texttt{ARTIFACT}  \\
    \bottomrule
    \end{tabularx}%
    }
    \caption{\textbf{Definition of ESCO Labels.} Indicated are the definitions of the ESCO labels used in this work taken from the ESCO taxonomy. Artifacts of the ESCO API are \texttt{K?} and \texttt{S?}, and \texttt{0000}, this means that no component was found.}
    \label{tab:lab}
\end{table*}
\clearpage

\begin{figure}[t]
    \centering
    \includegraphics[width=\linewidth]{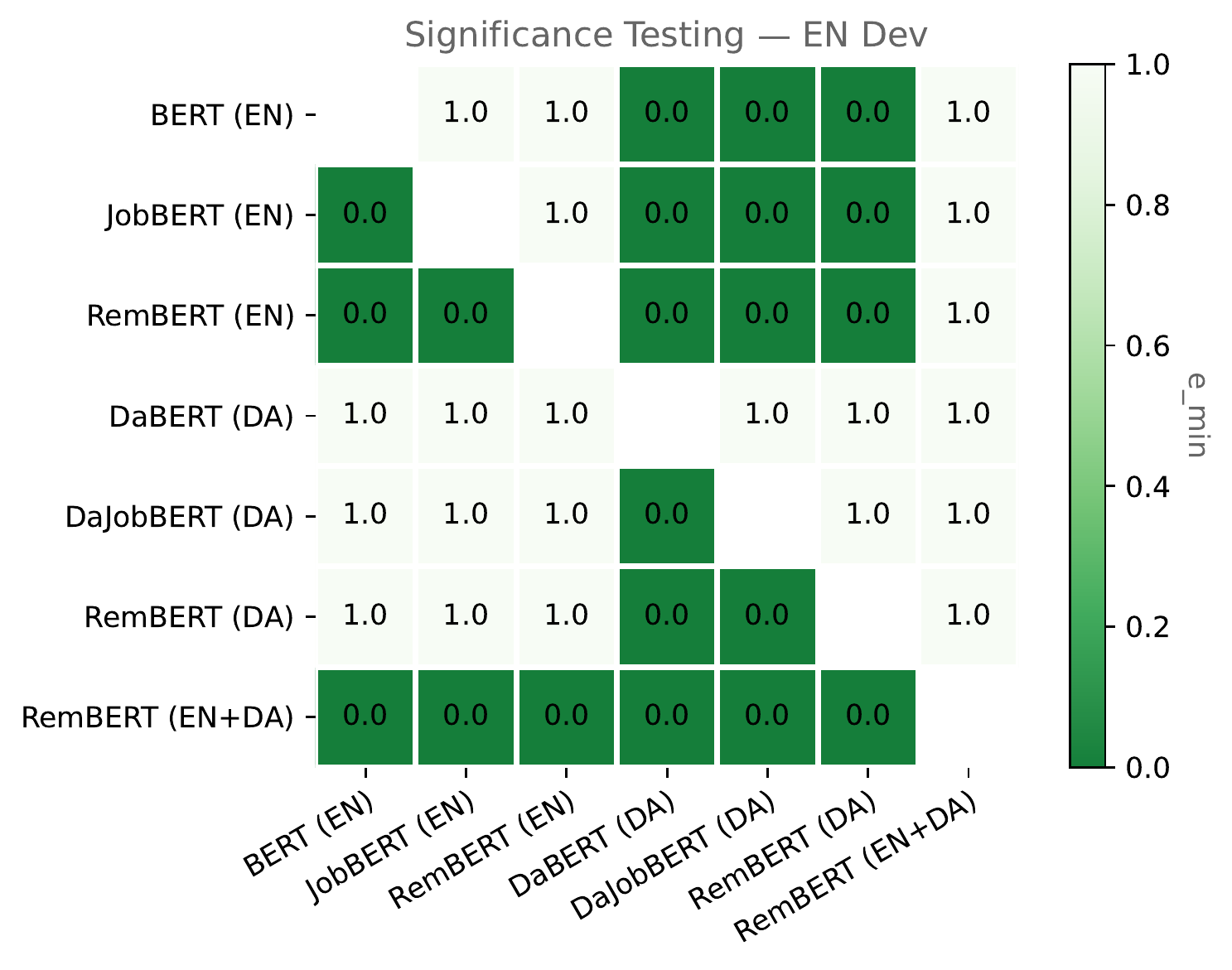}
    \includegraphics[width=\linewidth]{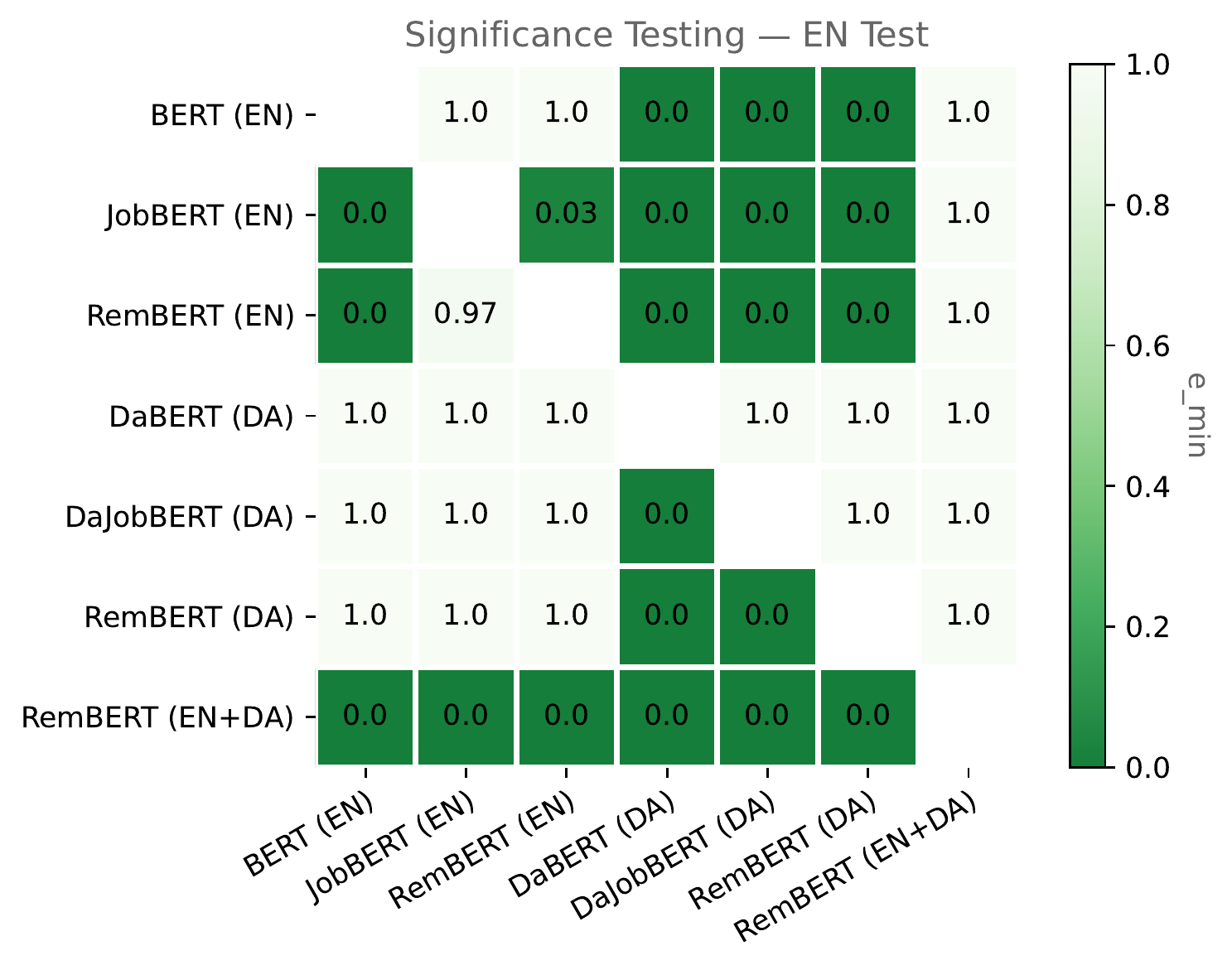}
    \includegraphics[width=\linewidth]{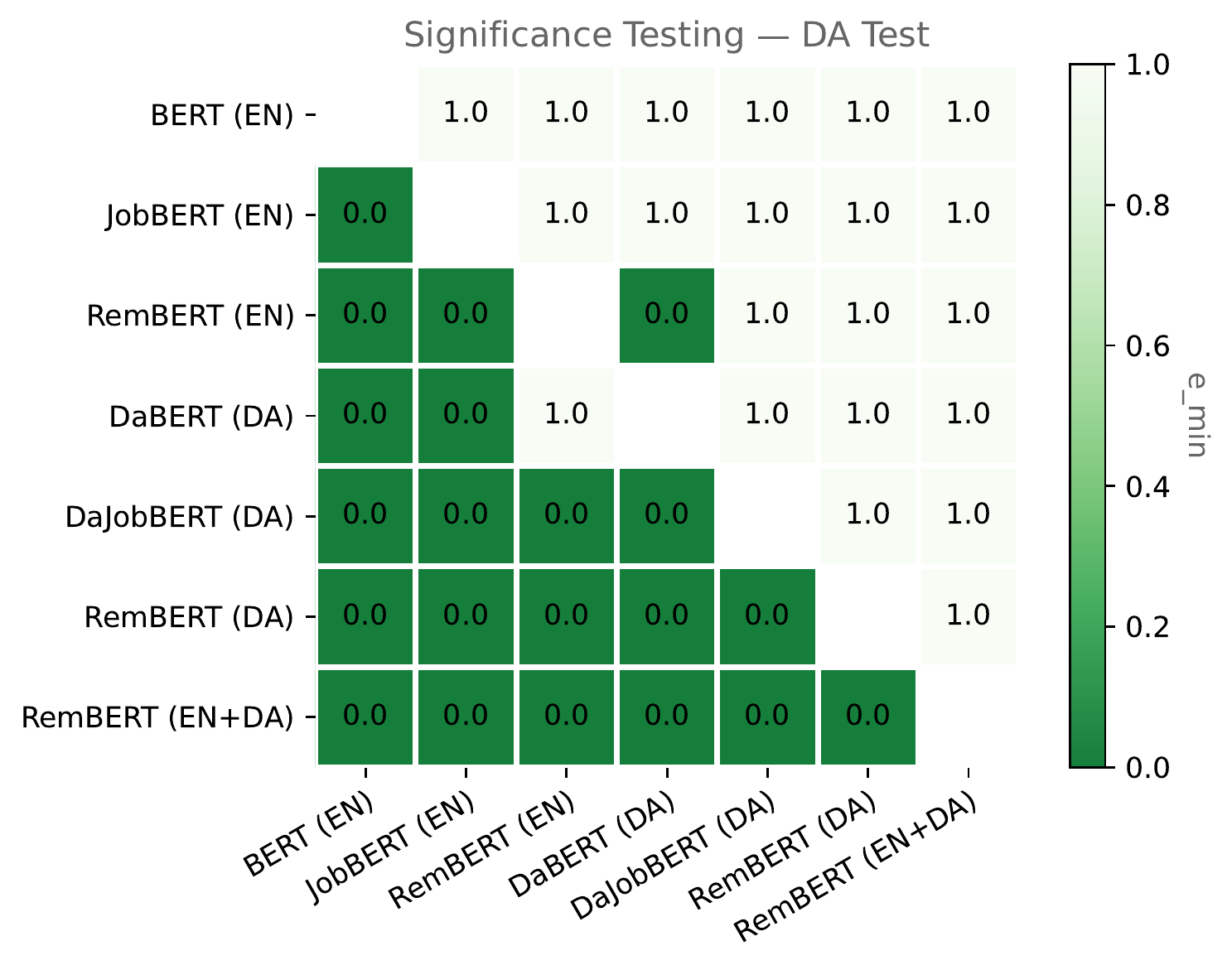}
    \caption{\textbf{Results Almost Stochastic Order.}
    ASO scores expressed in $\epsilon_\text{min}$.
    The significance level $\alpha =$ 0.05 is adjusted accordingly by using the Bonferroni correction~\protect\cite{bonferroni1936teoria}. Almost stochastic dominance ($\epsilon_\text{min} < 0.5$) is indicated in the colored boxes: On \textsc{\textbf{EN Test}}, JobBERT is almost stochastically dominant over RemBERT (EN), with $\epsilon_\text{min} = 0.03$.}
    \label{fig:aso_test}
\end{figure}

\begin{table}[t]
    \centering
    \resizebox{\linewidth}{!}{
    \begin{tabular}{l|r|r|r}
    \toprule
    \textsc{\textbf{Model}} & \textsc{\textbf{EN Dev}} & \textsc{\textbf{EN Test}} & \textsc{\textbf{DA Test}} \\
    \midrule
    \bertb{} (EN)        & \std{0.628}{0.004}  & \std{0.632}{0.007}     & \std{0.038}{0.008}\\
    JobBERT (EN)         & \std{0.628}{0.006}  & \textbf{\std{0.644}{0.006}*}     & \std{0.063}{0.005}\\
    RemBERT (EN)         & \textbf{\std{0.629}{0.003}}  & \std{0.637}{0.007}     & \std{0.354}{0.021}\\
    \midrule
    DaBERT (DA)         & \std{0.088}{0.013}  & \std{0.076}{0.012}      & \std{0.199}{0.058}\\
    DaJobBERT (DA)      & \std{0.101}{0.024}  & \std{0.096}{0.024}      & \std{0.395}{0.021}\\
    RemBERT (DA)        & \std{0.116}{0.052}  & \std{0.098}{0.040}      & \std{0.166}{0.141}\\
    RemBERT (EN+DA)     & \textbf{\std{0.629}{0.006}*}  & \std{0.643}{0.006}      & \textbf{\std{0.472}{0.014}*}\\
    \bottomrule
    \end{tabular}}
    \caption{\textbf{Exact Results on Splits.} Indicated are the exact results of the bar plots in~\cref{fig:results}. Significance tested with Almost Stochastic Order~\protect\cite{dror2019deep} test with Bonferroni correction~\protect\cite{bonferroni1936teoria}. Bold indicates highest average weighted macro-F1 and asterisk indicates significance.}
    \label{tab:exactresults}
\end{table}

\section{Exact Results from Plots}\label{app:results}

In~\cref{tab:exactresults}, we show the exact results of the plots from~\cref{fig:results} on English dev, English test, and Danish test respectively. In addition, we do significance testing. Recently, the Almost Stochastic Order (ASO) test~\cite{dror2019deep}\footnote{Implementation of~\newcite{dror2019deep} can be found at~\url{https://github.com/Kaleidophon/deep-significance}~\cite{dennis_ulmer_2021_4638709}} has been proposed to test statistical significance for deep neural networks over multiple runs.
Generally, the ASO test determines whether a stochastic order~\cite{reimers2018comparing} exists between two models or algorithms based on their respective sets of evaluation scores. Given the single model scores over multiple random seeds of two algorithms $\mathcal{A}$ and $\mathcal{B}$, the method computes a test-specific value ($\epsilon_\text{min}$) that indicates how far algorithm $\mathcal{A}$ is from being significantly better than algorithm $\mathcal{B}$. When distance $\epsilon_\text{min} = 0.0$, one can claim that $\mathcal{A}$ stochastically dominant over $\mathcal{B}$ with a predefined significance level. When $\epsilon_\text{min} < 0.5$ one can say $\mathcal{A} \succeq \mathcal{B}$. On the contrary, when we have $\epsilon_\text{min} = 1.0$, this means $\mathcal{B} \succeq \mathcal{A}$. For $\epsilon_\text{min} = 0.5$, no order can be determined. We compared all pairs of models based on five random seeds each using ASO with a confidence level of $\alpha =$ 0.05 (before adjusting for all pair-wise comparisons using the Bonferroni correction~\cite{bonferroni1936teoria}). Almost stochastic dominance ($\epsilon_\text{min} < 0.5$) is indicated in~\cref{fig:aso_test} over all the splits.

\begin{figure}[ht]
    \centering
        \includegraphics[width=\linewidth]{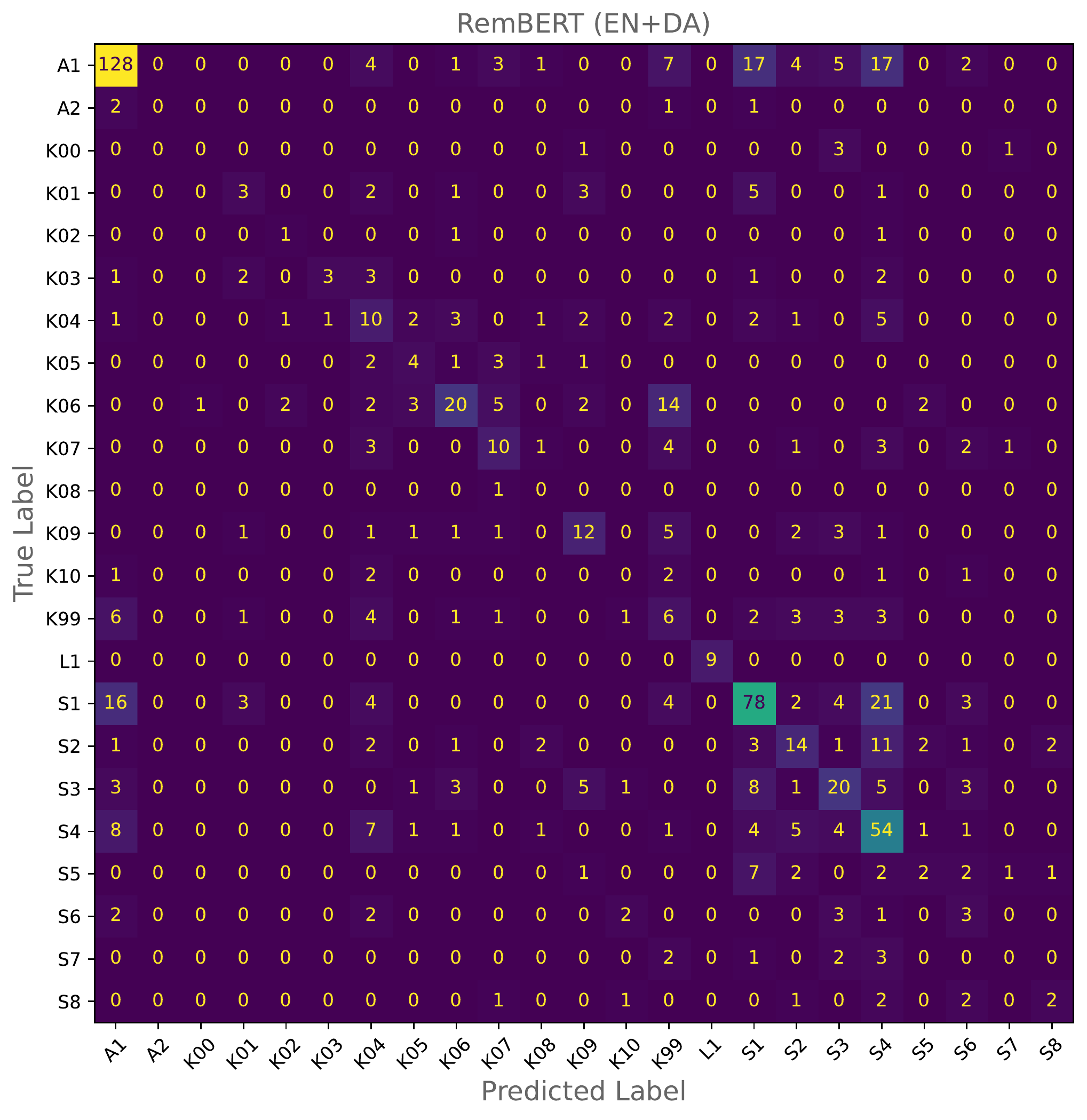}
    \caption{\textbf{Confusion Matrix of RemBERT (EN+DA).} We show the confusion matrix of the few-shot setting with RemBERT (EN+DA). On the diagonal are the correctly predicted labels. There is less confusion in this model as compared to RemBERT (EN). We suspect the additional Danish data benefits the prediction of \texttt{A1}.}
    \label{fig:cm}
\end{figure}

\section{Confusion Matrix Few-Shot}

In~\cref{fig:cm}, we show the confusion matrix of the best performing few-shot model on the test set of the best run and investigate what the model does not predict correctly. Dissimilar from \cref{fig:cm-en}, we only seem some confusion in the small cluster of \texttt{S1-4}. Giving the model a few Danish JPs substiantially improved the prediction of \texttt{A1}, which relates to \emph{attitudes} and gets predicted as \texttt{S1}: Communication, collaboration and creativity.

\end{document}